\documentclass[10pt,twocolumn,letterpaper]{article}

\usepackage{iccv}
\usepackage{times}
\usepackage{epsfig}
\usepackage{graphicx}
\usepackage{amsmath}
\usepackage{amssymb}

\usepackage{amsthm}

\usepackage{multirow}
\usepackage{floatrow}
\usepackage{anyfontsize}
\usepackage{sidecap}
\usepackage{floatrow}
\usepackage{graphicx}
\usepackage{amsmath}
\usepackage{amssymb}
\usepackage{booktabs}
\usepackage{xcolor}
\usepackage{soul}
\usepackage{colortbl}
\usepackage{algpseudocode,algorithm,algorithmicx}
\definecolor{Seashell}{RGB}{225, 225, 225} %背景色浅一点的
\definecolor{SeashellO}{RGB}{225, 180, 225} %背景色浅一点的
\definecolor{SeashellG}{RGB}{180, 225, 225} %背景色浅一点的
\definecolor{Seashellman}{RGB}{225, 120, 60} %背景色浅一点的
\definecolor{Seashellhorse}{RGB}{120, 60, 225} %背景色浅一点的
\definecolor{Seashellcow}{RGB}{0, 120, 120} %背景色浅一点的
\definecolor{Firebrick4}{RGB}{0, 0, 0}%文字颜色红一点的

\newcommand{\code}[1]{
	\begingroup
	\sethlcolor{Seashell}%背景色
	\textcolor{Firebrick4}{\hl{#1}}%textcolor里面对应文字颜色
	\endgroup
}

\newcommand{\zlorange}[1]{
	\begingroup
	\sethlcolor{SeashellO}%背景色
	\textcolor{Firebrick4}{\hl{#1}}%textcolor里面对应文字颜色
	\endgroup
}

\newcommand{\zlgreen}[1]{
	\begingroup
	\sethlcolor{SeashellG}%背景色
	\textcolor{Firebrick4}{\hl{#1}}%textcolor里面对应文字颜色
	\endgroup
}
\newtheorem{prop}{Proposition}[]

\usepackage[pagebackref=true,breaklinks=true,letterpaper=true,colorlinks,bookmarks=false]{hyperref}

\iccvfinalcopy % *** Uncomment this line for the final submission

\def\iccvPaperID{2334} % *** Enter the ICCV Paper ID here

\ificcvfinal\fi

\begin{document}

%%%%%%%%% TITLE
\title{A Retrospect to Multi-prompt Learning across Vision and Language}

\author{
	Ziliang Chen$^{1,3}$, \ Xin Huang$^{2}$, \ Quanlong Guan$^{1}$\thanks{indicate corresponding author. }, \ Liang Lin$^{2}$ \ Weiqi Luo$^{1}$\\ \begin{small}
		$^1$Jinan University \ \ $^2$Sun Yat-sen University \ \ $^3$Pazhou Laboratory 
	\end{small}\\
	\tt\scriptsize c.ziliang@yahoo.com, \tt\scriptsize huangx353@mail2.sysu.edu.cn, \tt\scriptsize \{Gql,lwq\}@jnu.edu.cn, \tt\scriptsize linliang@ieee.org 
	% For a paper whose authors are all at the same institution,
	% omit the following lines up until the closing ``''.
	% Additional authors and addresses can be added with ``\and'',
	% just like the second author.
	% To save space, use either the email address or home page, not both
}

\iffalse\author{Ziliang Chen\\
Jinan University\\
{\tt\small c.ziliang@yahoo.com}
% For a paper whose authors are all at the same institution,
% omit the following lines up until the closing ``}''.
% Additional authors and addresses can be added with ``\and'',
% just like the second author.
% To save space, use either the email address or home page, not both
\and
Xin Huang\\
Sun Yat-sen University\\
{\tt\small huangx353@mail2.sysu.edu.cn}
\and
Quanlong Guan\\
Jinan University\\
{\tt\small Gql@jnu.edu.cn}
\and
%Liang Lin\\
%Sun Yat-sen University\\
%{\tt\small secondauthor@i2.org}
%\and
%Weiqi Luo\\
%Jinan University\\
%{\tt\small secondauthor@i2.org}
}\fi

\maketitle
% Remove page # from the first page of camera-ready.
\ificcvfinal\thispagestyle{empty}\fi

\begin{abstract}
	The vision community is undergoing the unprecedented progress with the emergence of Vision-Language Pretraining Models (VLMs). Prompt learning plays as the holy grail of accessing VLMs since it enables their fast adaptation to downstream tasks with limited resources. Whereas existing researches milling around single-prompt paradigms, rarely investigate the technical potential behind their multi-prompt learning counterparts. This paper aims to provide a principled retrospect for vision-language multi-prompt learning. We extend the recent constant modality gap phenomenon to learnable prompts and then, justify the superiority of vision-language transfer with multi-prompt augmentation, empirically and theoretically. In terms of this observation, we propose an Energy-based Multi-prompt Learning (EMPL) to generate multiple prompt embeddings by drawing instances from an energy-based distribution, which is implicitly defined by VLMs. So our EMPL is not only parameter-efficient but also rigorously lead to the balance between in-domain and out-of-domain open-vocabulary generalization. Comprehensive experiments have been conducted to justify our claims and the excellence of EMPL.   
\end{abstract}
%%%%%%%%% ABSTRACT
\section{Introduction}\vspace{-4pt}
\label{sec:intro}

Recent years have witnessed the rise of multimodal intelligence, in particular, Vision-Language Pre-training models (VLMs), \emph{e.g.}, CLIP \cite{CLIP}, ALIGN \cite{ALIGN}, achieving downstream tasks in low resources by converting the prior knowledge behind large language models (LLMs) \cite{devlin2018bert,NEURIPS2020_GPT3}. {Given a pair of image encoder (\emph{e.g.}, ResNet \cite{he2016deep}, ViT \cite{dosovitskiy2020image}, \emph{etc}) and text encoder (\emph{i.e.}, LLMs), VLMs align visual features with their corresponding textual description embeddings via contrastive learning \cite{du2022learning,he2020momentum,zolfaghari2021crossclr}. So provided a text known as \emph{prompt}, {VLMs may rapidly adapt to diverse tasks \cite{du2022learning,luddecke2022image} by matching harmony visual patterns with the textual description}. 
%	(\emph{e.g.}, ``\texttt{a photo of a \{class\}.}'' used for image classification). 
	The principle sheds a new light in computer vision for in-domain and out-of-domain generalization.
	
		\begin{figure}[t]
		\center
		\includegraphics[width = 0.9\columnwidth]{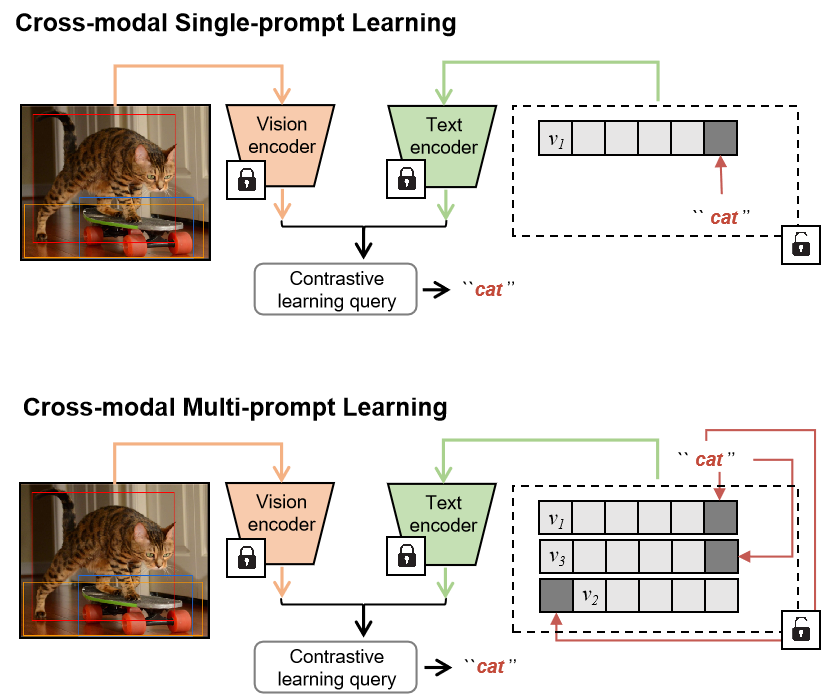}
		\vspace{-2pt}
		\caption{The overview of cross-modal single-prompt learning and multi-prompt learning (MPL). With more prompt templates, MPL brings new opportunities and challenges as discussed in the community, yet seldom giving a systematic investigation and solution.  
		}\vspace{-4pt}
		\label{fig1}
	\end{figure}

	The impressive cross-modal transferability behind VLM typically owes to the problem-customized prompting style, yet
	%\footnote{For instance, a prompt ``\texttt{a photo of a \{class\}.}'' is ideal for classfying a dog yet probably fail to categorizing a brambling since the latter tasks ask for more fine-grained decriptions pointing to its charateristics apart from other birds.}
	 demanding a great magnitude of trials and errors for selecting the ideal prompt template from a pool of candidates. Tedious workloads are consumed and do not guarantee the optimal prompt template either. Instead of the prompt engineering, \emph{prompt learning / tuning} \cite{CoOp} sidesteps the obstacle using soft prompting: outside of the words related with what we are interested in, the rest of textual slots in the template are replaced by a sequence of learnable context vectors ahead of the text encoder. In this principle, the optimal prompt template could be achieved by fine-tuning the learnable context vectors along with the given textual semantic while keeping the rest parameters of VLMs frozen for the optimization. {The data-driven merit increasingly arouses a flood of interests in the community \cite{CoCoOp,rao2022denseclip,sung2022vl}.} 
	
	Despite the significant progress, existing work of prompt learning focused on a single template whereas multi-prompt learnable context templates remain under-explored (Fig.\ref{fig1}). Recent NLP advance argue that instructing LLMs via more prompts may trigger its underlying in-context learning ability to master new skills \cite{NEURIPS2020_GPT3}. In this regard, multi-prompting is also deemed to be a promising trend for VLMs. On the other hand, existing studies remain confused of how multiple prompts work for VLMs, particularly from two aspects.  	
%	 where the cross-modal transferability have been scarcely considered in the LLM literature. 
 \hspace{-1pt}The first is \emph{\textbf{vision-language transferrability}} \cite{zhangdiagnosing}. Prompt augmentation eliminates the average cross-modal transfer shift \hspace{-0em}while \cite{CoCoOp} showed that increasing the scale of context vector tokens resulted in the detrimental effect, implying a larger cross-modal disparity. \hspace{-0.6pt}The second \hspace{-0.3pt}is\hspace{-0.3pt} {\emph{\textbf{open-vocabulary (OV) generalization}}, \emph{i.e.}, the model awareness of unseen classes.} Different from multple textual descriptions, learning with more prompts suggests more parameters. It casted a doubt of overfitting to training classes and put unseen classes at the risk of model generalization \cite{chen2022prompt}.

%	Existing approaches \cite{Alpher02,} almost skip the evaluation to wider unseen classes, thus, 
	
	In this paper, we provided several principled insights to understand multi-prompt learning \emph{empirically} and \emph{theoretically}. We first consider the cross-modal embedding space following the constant modality-gap phenomenon found by \cite{zhangdiagnosing}. With regards to our empirical observations, we extend the conclusion to learnable prompts to show that more learnable prompts might reduce the constant modality gap more significantly. In terms of constant modality gaps, we further proved the existence of \emph{\textbf{cross-modal unidentifiability issue}}: a paradox confusing the cross-modal model with a single prompt template in visual recognition. It could be restrained by multi-prompting empirically, thus, interpreting why multi-prompt learning could outperform single prompt for the sake of vision-language transferrability.     
	
	In terms of our retrospect, the main challenge of multi-prompt learning refers to its generalizability. Derived from this concern, we propose a new methodology Energy-based Multi-Prompt Learning (EMPL) for striking the balance between in-domain generalization and open-vocabulary generalization abilities. EMPL implicitly defines an energy-based \cite{lecun2006tutorial} prompt distribution that simultaneously use image and prompt as the variable. \hspace{-0.2em}With this regard, our method could be rigorously treated as modeling the uncertainty to explore the image-prompt embedding pairs with concepts out of the training domains, whereas also well generalizes to examples belonging to in-domain classes. The prompts are iteratively generated via a stochastic Markov Chain Monte Carlo (MCMC) sampler \cite{welling2011bayesian}, which is parameter-efficient, sensitive of input knowledge from vision-text encoders, and more importantly, general enough to cooperate with existing prompt learning strategies to upgrade the performances.  Experiments are comprehensively conducted to validate our claims and the superiority of our approach.
	
%	\textbf{Our contributions} are basically summarized as: \textbf{1}).We provide a systematic analysis derived from the embedding geometry behind prompt learning models, empirically and theoretically, verify that multi-prompt learning outperforms single-prompt learning based upon the cross-modal transferability; \textbf{2}). We reframe multi-prompt learning as energy-based models to propose our methodology EMPL. It draws prompts from a energy-based prompt distribution, which is parameter-efficient, sensitive of adaptive information from encoders, and model-agnostic. We offer a theoretical explanation why it excels for in-domain and out-of-domain generalization.	\textbf{3}). Comprehensive experiments are conducted to validate our claims and the superiority of our method. 
	
{\color{red}	       

	}

	\vspace{-0pt}\section{Related Work}\vspace{-0pt}
	\label{sec:formatting}
	
	\textbf{Vision-Language Pre-trained (VLM) models.} VLM models, which unify the two most commonly used modalities, vision and language, have gained great popularity due to the success of pre-trained models \cite{NEURIPS2020_GPT3} in CV and NLP. Among numerous VLM models \cite{CLIP, ALIGN}, CLIP \cite{CLIP} is the most widely used and representative one. It utilizes a pair of image and text encoders to receive information from both modalities and leverages a large amount of paired image and text data collected from the Internet. In contrast to other VLM models that use Masked language modeling \cite{kim2021vilt, lu2019vilbert}, Masked region prediction \cite{tan2019lxmert, su2019vl}, etc., CLIP utilizes feature vectors from both encoders to train with the Contrastive Learning \cite{CLIP} strategy, which successfully aligns the feature space of both modalities and has been widely employed for a variety of downstream tasks \cite{gu2021open, ghiasi2021open, nichol2021glide}. 

\textbf{Prompt learning.} Prompt tuning \cite{petroni2019language, radford2019language}, a technique derived from the field of natural language processing (NLP), has gained great popularity \cite{CoOp,CoCoOp,lu2022prompt} in the field of VLP in recent years, which has the ability to unleash the potential of pre-trained multimodal models. CoOp \cite{CoOp} , a well-known text branching technique, eliminates the need for manual prompt design by transforming input context tokens into learnable vectors. CoCoOp \cite{CoCoOp}, its successor, overcomes its generalization issues by taking visual features into account when creating prompts. Bahng et al.\cite{bahng2022exploring} propose a visual prompt approach by adding task-specific, learnable visual signals into images. Additionally, prompt learning has been employed to equip VLP models with the ability to tackle a variety of tasks, including open-vocabulary object detection \cite{du2022learning}, semantic segmentation \cite{luddecke2022image, rao2022denseclip}, and scene graph generation \cite{he2022towards}.

		\textbf{Multi-prompt learning.} More recently, the advantages of building multiple context templates for prompt learning have been empirically verified. For instance, \cite{lu2022prompt,derakhshani2022variational} provided a distributional point of view to model the learnable template, in which the diversity across templates were emphasized; \cite{chen2022prompt} trained multiple prompts by decreasing the cross-modality optimal transport \cite{peyre2019computational,chen2020graph} across the prompt embeddings and visual embeddings to match different visual aspects by different templates; \cite{ge2022domain} learns different prompts to specify domain information, \emph{etc}. These work demonstrate the promising outlook of multi-prompt learning for VLMs, {whereas their solutions are typically heurstic and specific, limited to inspire the research in this thread.} 
	%\textbf{Gromov-Wasserstein (GW) distance.}
	
	%-------------------------------------------------------------------------

	%------------------------------------------------------------------------

%	In this section, we first provide the reviews on CLIP and some classical methods for prompt learning based on CLIP. Afterwards, we elaborate the motivation and technical details of GAPL as well as the	inference and optimization rationale behind our approach. 
	
	\vspace{-0pt}\section{Background}\vspace{-0pt}
	Here we give a brief review of cross-modal prompt engineering and learning, then generalize the notations to multi-prompt strategies prepared for the elaboration of our work.
	
	\textbf{Contrastive Language-Image Pre-training.} CLIP consists of a pair of visual encoder $f$ and text encoder $h$, which take ResNet / ViT and BERT \cite{devlin2018bert} as their backbones. \hspace{-0.2em}Given an image $\boldsymbol{x}$ with its label \hspace{-0.2em}${c}$ contained in the description $\boldsymbol{y}(c)$, 
	CLIP extracts a feature $f$$=$$f(\boldsymbol{x})$ by the visual encoder, then taking the text encoder $h(\cdot)$ to align $f$ with the text embedding $\boldsymbol{h}_{c}$$=$$h\big(\boldsymbol{y}(c)\big)$ generated from $\boldsymbol{y}(c)$, \emph{e.g.}, ``\texttt{a cropped photo of }\{$c$\}''. $f$ and $h$ are trained with tons of image-caption pairs to bridge the modalities by contrastive representation learning based on the prediction probability:
	\vspace{-0pt}\begin{equation}\label{clip}
		\begin{aligned}
			P(\texttt{class}=c|\boldsymbol{x})=\frac{\exp\big({\rm{sim}}(f, \boldsymbol{h}_{c})/\gamma\big)}{\sum_{i=1}^{K}\exp\big({\rm{sim}}(f, \boldsymbol{h}_{c_{i}})/\gamma\big)},
		\end{aligned}
	\end{equation}\vspace{-0pt}where $c$ is supposed to classify into the $K$ classes $\{c_{i}\}^K_{i=1}$; ${\rm{sim}}(\cdot,\cdot)$ denotes a metric function such as cosine similarity, and $\gamma$ denotes the temperature
	of Softmax. 
	
	\textbf{Prompt Learning.} The template $\boldsymbol{y}(\cdot)$ is engineered before matching $f$ and $\boldsymbol{h}_{c}$ by CLIP. Instead, CoOp \cite{CoOp} learns a set of vectors $\boldsymbol{v}$$=$$\{\boldsymbol{v}_1,\boldsymbol{v}_2,\cdots,\boldsymbol{v}_m\}$ to replace $\boldsymbol{y}(\cdot)$, each of which was understood as a pseudo word embedding and $m$ denotes the length of the learnable word slots. So prompting becomes $\boldsymbol{h}_{\boldsymbol{v}}(c)$$=$$\{\boldsymbol{v}_1,\boldsymbol{v}_2,\cdots,\boldsymbol{v}_m, \boldsymbol{v}(c)\}$ where $\boldsymbol{v}(c)$ denotes the embedding of the class name of $c$. Therefore given a pre-trained CLIP, CoOp tunes the prompt parameter $\boldsymbol{v}$ to achieve a downstream task with a frozen CLIP:
	\begin{equation}\label{coop}
		\begin{aligned}
			P_{\boldsymbol{v}}(\texttt{class}\hspace{-0.2em}=\hspace{-0.2em}y|\boldsymbol{x})=\frac{\exp\big({\rm{sim}}(f, \boldsymbol{h}_{\boldsymbol{v}}(c))/\gamma\big)}{\sum_{i=1}^{K}\exp\big({\rm{sim}}(f, \boldsymbol{h}_{\boldsymbol{v}}(c_i))/\gamma\big)},
		\end{aligned}
	\end{equation}in which $\boldsymbol{h}_{\boldsymbol{v}}(c)$ can be specified to represent a wider range of prompt learning paradigms \cite{CoCoOp,rao2022denseclip}.    
	
	\textbf{Generic Notations of Multi-prompt Learning.} Most existing work\hspace{-0.1em} of multi-prompt engineering\hspace{-0.1em} and learning are derived from Eq.\ref{clip} and Eq.\ref{coop}, therefore we may extend their notations to generally represent the multi-prompt methods. In particular, we employ $\boldsymbol{H}\big(c;\mathcal{V}\big)$ instead of $\boldsymbol{h}_{c}$ to denote a set of prompts composed of a vocabulary $\mathcal{V}$ and contains the word $c$. CLIP-derived prompt engineering approaches can be generally concluded into:
	\vspace{-0pt}\begin{equation}\label{mpe}
	\begin{aligned}
	P(\boldsymbol{x})[c]=\frac{\exp\big({\rm{sim}}(f(\boldsymbol{x}), \boldsymbol{H}(c;\mathcal{V}))/\gamma\big)}{\sum_{i=1}^{K}\exp\big({\rm{sim}}(f(\boldsymbol{x}), \boldsymbol{H}(c_i;\mathcal{V}))/\gamma\big)},
	\end{aligned}
	\end{equation}\vspace{-0pt}where ${\rm{sim}}(\cdot,\cdot)$ denotes a generic metric to estimate the difference between the feature $f(\boldsymbol{x})$ and the multi-prompt embeddings $\boldsymbol{H}(c;\mathcal{V})$. If $\boldsymbol{H}(c;\mathcal{V})$ can be learned, we use $\boldsymbol{\boldsymbol{\phi}}$$=$$\{\boldsymbol{\theta},\boldsymbol{v}\}$ to denote the context $\boldsymbol{v}$ and other learnable parameters $\boldsymbol{\theta}$, then rewrite Eq.\ref{mpe} by $\boldsymbol{H}_{\boldsymbol{\phi}}(c;\mathcal{V})$:     
	\vspace{-0pt}\begin{equation}\label{mpl}
	\begin{aligned}
	P_{\phi}(\boldsymbol{x})[c]\hspace{-0.2em}=\hspace{-0.2em}\frac{\exp\big({\rm{sim}}(f(\boldsymbol{x}), \boldsymbol{H}_{\boldsymbol{\phi}}(c;\mathcal{V}))/\gamma\big)}{\sum_{i=1}^{K}\exp\big({\rm{sim}}(f(\boldsymbol{x}), \boldsymbol{H}_{\boldsymbol{\phi}}(c_i;\mathcal{V}))/\gamma\big)}.
	\end{aligned}
	\end{equation}In terms of task goals, the vocabulary $\mathcal{V}$ in Eq.\ref{mpl} is only interested in the words related with the task, which has already summed up a set of multi-prompt learning methods \cite{lu2022prompt,chen2020graph}. 

\vspace{-0pt}\section{Embedding Geometry behind Prompts}\vspace{-0pt}
	The core of CLIP-derived prompt learning hinges on the prompt template's ability to solve vision tasks with text embeddings as inquiry proxies. The cross-modal transferability is achieved for a pair of an image $\boldsymbol{x}$ and its description $\boldsymbol{y}(c)$ if a vision classifier outputs similar predictions on their embeddings. While given a matched image-text pair, the embeddings extracted by CLIP counter-intuitively persisted a modality gap {\cite{zhangdiagnosing,liang2022mind}}. Our work demonstrated this geometrical phenomenon also widely exist in prompt learning with virtual description $\boldsymbol{v}(c)$ and results in \emph{cross-modal non-identifiability issues} in single-prompt learning. {To this, the technical merit of multi-prompt strategies can be verified from the view of cross-modal transferrability.} 
	
	\textbf{Modality Gaps between Images and Prompts.} As proposed in \cite{liang2022mind}, the modality gap is caused by contrastive optimization and can be categorized into two types: the \emph{individual}-level modality gap $\boldsymbol{g}(\boldsymbol{x},\boldsymbol{y})$ differentiates the embeddings for a image-text pair $(\boldsymbol{x},\boldsymbol{y})$; the \emph{class}-level modality gap $\boldsymbol{g}(c)$ differentiates the average between the embeddings for images and text related with the class $c$, namely, 
	\vspace{-0pt}\begin{equation}\label{gap}
	\begin{aligned}
	&\boldsymbol{g}(\boldsymbol{x},\boldsymbol{y}) &=&f(\boldsymbol{x})\hspace{-0.2em}-\hspace{-0.2em}h(\boldsymbol{y}) ,\ \ \ \ \ \ \forall \big(\boldsymbol{x}, \boldsymbol{y}\big)\sim P_{\mathcal{X}\times\mathcal{Y}};\\
	&\boldsymbol{g}(c) &=&\mathbb{E}_{\boldsymbol{x}\sim P_{\mathcal{X}|c}} f(\boldsymbol{x})\hspace{-0.2em}-\hspace{-0.2em}\mathbb{E}_{\boldsymbol{y}\sim P_{\mathcal{Y}|c}}h(\boldsymbol{y}).
	\end{aligned}
	\end{equation}Observed across a range of contrastive multimodal models, \emph{the modality gaps $\boldsymbol{g}(\boldsymbol{x},\boldsymbol{y})$ and $\boldsymbol{g}_c$ could be approximated by a constant vector} \cite{zhangdiagnosing}. These embedding geometrical properties provide a new explanation why prompt learning can outperform CLIP: learning to prompt might implicitly reduce the modality gap constant between image-text pairs.   

	To procure the evidences of our guessing, we investigate the embedding geometry derived from the prompt embeddings extracted from the single-prompt learner CoOp \cite{CoOp} and multi-prompt learner ProDA \cite{lu2022prompt}. The means and variances of the magnitude ($|\boldsymbol{g}|$) and direction ($\cos(\boldsymbol{g},\mathbb{E}_{\boldsymbol{g}}\boldsymbol{g})$) are estimated to justify whether the individual and class modality gaps can be approximated by a constant vector\footnote{We follows the same evaluation setup in \cite{zhangdiagnosing}.}. In Fig.\ref{mg}, CoOp and three ProDA variant models with different scale of prompts preserve the average gaps with trivial variances in their magnitudes and directions, thus, their modality gaps have been approximated by some constant vectors, respectively. On account of this observation, prompt learner differ in their modality gap magnitudes: CoOp trained from CLIP can further minimize the gap magnitude, however, underperforms ProDA with the prompt augmentation strategy: adding prompts notably leads to closing the modality gaps. Hence \emph{the failure of the overextended scale of prompts} \emph{more likely results from the overfitted model rather than the incompetence of bridging the modality disparity.}    
	
		\begin{figure}[t]
		\center
		\includegraphics[width = 1\columnwidth]{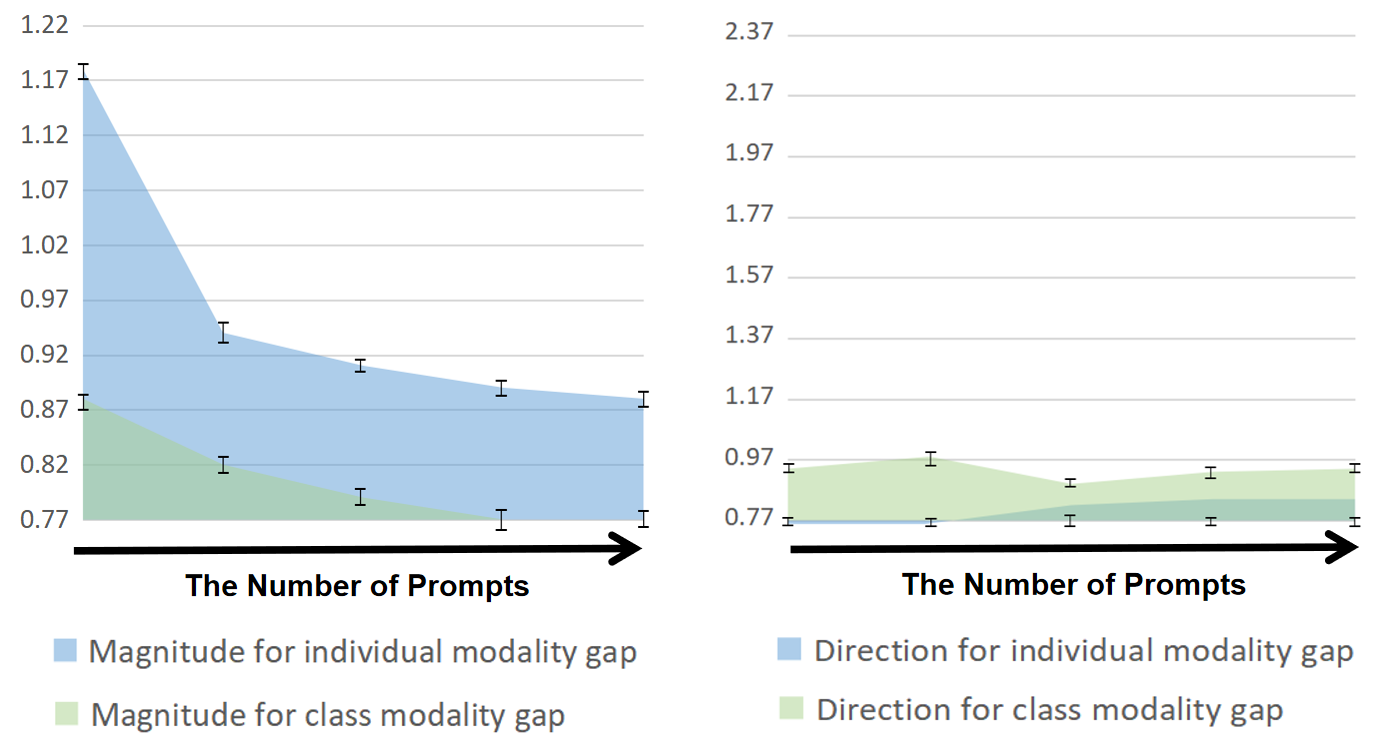}
		\vspace{-8pt}
		\caption{Magnitude (\textbf{M}) and Direction (\textbf{D}) of individual modality gap (IMG) and class modality gap (CMG) on MsCOCO \cite{lin2014microsoft}. We gradually increase the number of prompts by switching models  as CLIP$\rightarrow$CoOp$\rightarrow$ProDA$\rightarrow$ProDA(x2)$\rightarrow$ProDA(x4), to observe the change of IMG and CMG.
		}\vspace{-10pt}
		\label{mg}
	\end{figure}
	 
\iffalse		\begin{table}[]
			\centering
			\setlength{\tabcolsep}{2.5pt}
			{\fontsize{7}{9.9}\selectfont
				\begin{tabular}{cc| c c |c cc}
					\hline
					\multicolumn{2}{c|}{\textbf{Baselines}} &CLIP &CoOp &ProDA(base) &ProDA($\times$2) &ProDA($\times$4)  \cr
					& &\multicolumn{2}{c}{{\textbf{\emph{Single-prompt method}}}} &\multicolumn{3}{|c}{\textbf{\emph{Multi-prompt Learning}}}  \cr
					\rowcolor{Seashell}\multirow{2}*{IMG}
					&\textbf{M} &{1.18${{\pm0.03}}$} &{0.94${{\pm0.04}}$} &{0.91${{\pm0.02}}$} &{0.89${{\pm0.03}}$}  &{\textbf{0.88}${{\pm0.02}}$}   \\
					&\textbf{D} &{{0.7}${{\pm0.06}}$} &0.76${{\pm0.01}}$ &{0.82${{\pm0.06}}$} &0.84${{{\pm0.03}}}$ &0.84${{{\pm0.02}}}$  \\
					\rowcolor{Seashell}\multirow{2}*{CMG}
					&\textbf{M} &{0.88${{\pm0.04}}$} &{0.82${{\pm0.05}}$} &{0.79${{\pm0.03}}$} &0.77${{\pm0.03}}$  &{\textbf{0.77}${{\pm0.04}}$} \\
					&\textbf{D} &{0.94${{\pm0.04}}$} &{0.98${{\pm0.03}}$} &{0.89${{\pm0.05}}$} &0.93${{\pm0.03}}$ &0.94${{\pm0.04}}$ \\
					\hline
				\end{tabular}}\vspace{5pt}\caption{Magnitude (\textbf{M}) and Direction (\textbf{D}) of individual modality gap (IMG) and class modality gap (CMG) on MsCOCO \cite{lin2014microsoft}. }\label{mg1}\vspace{-16pt}
			\end{table} \fi
			
	\textbf{Cross-modal Non-identifiability Issue.} In terms of the modality \hspace{-0.2em}gap, there is a issue that might happen if we use a single template to prompt a cross-modal contrastive model. \hspace{-0.5em}
	Concretely, let's consider two images $\boldsymbol{x}_i$, $\boldsymbol{x}_j$ with mutually exclusive concepts in visual realism, \emph{e.g.}, $\boldsymbol{x}_i$ belongs to $c_1$, $c_2$ and $\boldsymbol{x}_j$ belongs to $c_2$, $c_3$. Given a single prompt template $\boldsymbol{v}$ to convey these concepts, feature-embedding pairs $(f(\boldsymbol{x}_i)$, $h_{\boldsymbol{v}}(c_1))$, $(f(\boldsymbol{x}_i)$, $h_{\boldsymbol{v}}(c_2))$, and $(f(\boldsymbol{x}_j)$, $h_{\boldsymbol{v}}(c_2))$ are supposed to be approximated by the individual-level modality gap constant vector $\boldsymbol{c}$. Given this, we can prove $f(\boldsymbol{x}_j)$ far away from $h_{\boldsymbol{v}}(c_1)$ with the same constant vector $\boldsymbol{c}$,   
	\begin{prop}{\textbf{Individual-level cross-modal non-identifiability (Informal)}}
		Suppose a single-prompt learning model $\big(f(\cdot),h_{\boldsymbol{v}}(\cdot)\big)$ satisfies the constant individual-level modality gap. Given each pair of images $\boldsymbol{x}_1$, $\boldsymbol{x}_2$ with mutually exclusive concepts, it is not able to distinguish them by single-prompting with their exclusive concepts.
%		\begin{displaymath}
%		p_{\boldsymbol{\Phi}^\ast,\boldsymbol{\Theta}^\ast}(\boldsymbol{\hat{\boldsymbol{x}}}_j)=\int p_{\boldsymbol{\Phi}^\ast,\boldsymbol{\Theta}^\ast}(\boldsymbol{\hat{\boldsymbol{x}}}_j|\boldsymbol{x}_i)p_i(\boldsymbol{x}_i)d\boldsymbol{x}_i=p_{j}(\boldsymbol{\hat{x}}_j).
%		\end{displaymath}
	\end{prop}The formal statement and proof refer to our Appendix.A.
	Derived from the result, the image $\boldsymbol{x}_i$ and $\boldsymbol{x}_j$ can not be distinguished in terms of the proxy $\boldsymbol{v}(c_1)$, which should have been distinguished since the concept $c_1$ is exclusive for the image $\boldsymbol{x}_i$ in terms of the image $\boldsymbol{x}_j$. 
	
	Here we discussed a simple case for illustrating the issue: suppose we having a pair of \emph{{\color{Seashellhorse}horse}} (mutual concept) images, where the first image refers to the scene of \emph{a {\color{Seashellman}man}} (exclusive concept) \emph{riding a {\color{Seashellhorse}horse}} and the second describes that \emph{{\color{Seashellhorse}horses} and {\color{Seashellcow}cows}} (exclusive concept) \emph{drink nearby a river}. Prompting the images by the given captions is capable to differentiate the images in terms of their exlusive concepts \emph{{\color{Seashellman}man}} and \emph{{\color{Seashellcow}cow}}. However, given a single prompt only built with learnable context vectors in $\boldsymbol{v}$, inquiring with \emph{{\color{Seashellcow}cow}} or \emph{{\color{Seashellman}man}} is hard to classify these images if the context optimization satisfies the constant modality gap pressumption, in which the single template was optimized to overlook the other descriptive information beyond the classes (\emph{e.g.}, \emph{riding} and \emph{nearby river}). {The case is general since the images with mutually exclusive concepts may also refer to semantic information incorporated from the visual encoder \cite{CoCoOp}.} 
	
	What's worse, the class-level modality gap resembles the tragedy across different groups, arousing the chaos to identify embedding sets belonging to different concepts:   
	
	\begin{prop}{\textbf{Population-level cross-modal non-identifiability (Informal)}}
		Suppose a single-prompt learning model $\big(f(\cdot),h_{\boldsymbol{v}}(\cdot)\big)$ satisfies the constant individual-level modality gap. Given image groups $\boldsymbol{X}_1$, $\boldsymbol{X}_2$ have mutually exclusive concepts, it is not able to distinguish the groups via single-prompting with their group-specific concepts.
		%		\begin{displaymath}
		%		p_{\boldsymbol{\Phi}^\ast,\boldsymbol{\Theta}^\ast}(\boldsymbol{\hat{\boldsymbol{x}}}_j)=\int p_{\boldsymbol{\Phi}^\ast,\boldsymbol{\Theta}^\ast}(\boldsymbol{\hat{\boldsymbol{x}}}_j|\boldsymbol{x}_i)p_i(\boldsymbol{x}_i)d\boldsymbol{x}_i=p_{j}(\boldsymbol{\hat{x}}_j).
		%		\end{displaymath}
	\end{prop}The existence of non-identifiability validates \emph{the remarkably more effecting of multi-prompt learning strategy compared with single-prompt learning}: learning to prompt with multiple templates suggests remodeling the diversity of captions that could helpfully alleviate the issues. It is reflected by evaluating different prompt models on their prediction consistencies to images with mutually exclusive classes. Some evidences agreed with our conjecture in our empirical studies on Language-to-Image Retrivel and Multi-label classification (Appendix.C).

		\vspace{-0pt}\section{Energy-based Multi-prompt Learning} \vspace{-0pt}
		In the previous section, we discussed why multi-prompt learning benefits the cross-modal transferrability, though it does not reflect a model's generalization ability to adapt the cases beyond the prompt-tuning stage. The OV generalization is remarkable in CLIP yet it might rapidly deteriorate by prompt-tuning the backbone due to the traded-off performance on in-domain images. Subsequent techniques developed to resist the degeneration \cite{CoOp} were barely motivated by the multi-prompt learning charateristics. It is somehow because multi-prompt learning with more templates is supposed to bring more learnable context tokens, increasing the risk of overfitting.
	    Our concern rises from this regard: 
		
%		On the other side, . It encourages us to develop a 
		
		%To this end, we extend Meta-Net \cite{bibid} to the multi-prompt variant, \emph{i.e.}, \emph{multi-prompt Meta-Net} (M$^2$Net), which generates multiple templates for each image by encoding diverse visual features from $f(\boldsymbol{x})$. In this way, multi-prompt learning is more parsimonious in the parameter usage since the scale of templates is only determined by the size of $f(\boldsymbol{x})$.
	
	%In this manner, sapatial information can be pre-separated by $f(\boldsymbol{x})$ {to prevent images from sharing the prompt templates to inquire the mutual concept.}

%	Ought to be regarded that M$^2$Net is a strategy-agnostic method so that existing multi-prompt learning methods can be armed with M$^2$Net by replacing their multi-prompt generation pipelines. \hspace{-0.2em}As the multi-prompt strategy turns to rely on $\boldsymbol{x}$, we prefer $\boldsymbol{H}_{\boldsymbol{\phi}}\big(\mathcal{V}(c),\boldsymbol{x}\big)$ to $\boldsymbol{H}_{\boldsymbol{\phi}}(\mathcal{V}(c))$ in Eq.\ref{mpl} to denote the strategies supported by M$^2$Net. 

\vspace{-6pt}    \begin{center}
    	\emph{Is there a multi-prompt learning algorithm \\that simultaneously benefit the cross-modal transferrability (in-domain) and the OV generalization (out-domain)?}
    \end{center}\vspace{-6pt}
    
    To kill the two birds with one stone, we reinterpret multi-prompt learning from a persepective of energy-based models (\textbf{EBMs}) \cite{lecun2006tutorial}, where mutiple prompt templates are reproduced by drawing instances from an underlying EBM-based prompt distribution. The paradigm of Energy-based Multi-Prompt Learning (\textbf{EMPL}) is briefly shown in Fig.\ref{framework} and we further elaborate the methodology to demonstrate its potential to address our concern.    
    
     %also treated prompts as instances drawn from a prompt distribution, whereas our algorithm draws multiple prompts by reframing the objective into an . 
     
     %Distinct from existing prompt ditributions, our algorithm uses a more flexible density function and dynamically create prompts by .      
    
    \textbf{Energy-based Models.} The formulation of EBMs consists of an energy function $E(\cdot):$$\mathbb{R}^{D}$$\rightarrow \mathbb{R}$ that maps a $D$-dimensional datapoint into a scalar. It is implemented by a neural network with the parameter ${\theta}$, then $E_{\theta}(\cdot)$ is trained to assign the low energy to observed configurations of variables and deliver the high energy
    to unobserved ones. So given a dataset without the knowledge of its underlying density $p(Z)$, EBMs enable $p^{({\rm EBM})}_{\theta}(Z)=\frac{\exp(-E_{\theta}(Z))}{\int_{\boldsymbol{z}\in\mathcal{Z}}\exp(-E_{\theta}(\boldsymbol{z}))}$ to approximate the density function. The variable $Z$ in EBMs mostly refers to images or image features in previous work \cite{grathwohlyour,pang2020learning,wang2021energy}. In contrast, our EMPL depends on an energy function $E_{\phi}(Z)$$=$$E_{\phi}(X,H)$ with $Z$ constructed by the \emph{image variable} $X$ and the \emph{prompt variable} $H$. Therefore the multiple prompts in $\boldsymbol{H}_{\boldsymbol{\phi}}$ (Eq.\ref{mpl}) are obtained via drawing prompt instances from a EBM-based conditional prompt distribution $p^{({\rm EBM})}_{\phi}(H|X)$, \emph{i.e.},
    \vspace{-4pt}\begin{equation}\label{ebm}
    	\begin{aligned}
    		&\forall \ \boldsymbol{x} \sim p(X)=P_{\mathcal{X}}, %\ \ \exists \ \boldsymbol{h} \ \in \boldsymbol{H}_{\boldsymbol{v}};
    		\\
    		&\boldsymbol{H}_{\phi}\sim p^{({\rm EBM})}_{\phi}(H|\boldsymbol{x})=\frac{\exp(-E_{\phi}(\boldsymbol{x},H))}{\int_{H\sim\mathcal{H}}\exp(-E_{\phi}(\boldsymbol{x},H))},
    	\end{aligned}\vspace{-4pt}
    \end{equation}
    in which $\mathcal{H}$ denotes the space of soft prompts and the energy function is derived from the contrastive score $P_{\phi}(\cdot)$ in Eq.\ref{mpl}. %In this manner, the muliple prompt embeddings are  
    
        \textbf{Energy-based Open-vocabulary Learning.} Given an open vocabulary $\mathcal{V}$, we elaborate the meta-learning objective with the energy function $E_{\phi}(X,H)$ for improving in-domain and out-of-domain generalization. Instead of a $K$-word vocabulary in Eq.\ref{mpl}, the open vocabulary $\mathcal{V}$ demands a \emph{meta-classifier} of predicting arbitrary classes across visual recognition tasks. So EMPL is suggested to incorporate all the words in $\mathcal{V}$ and accordingly, $E_{\phi}(X,H)$ should support meta-learning to achieve a set of $K$-class visual recognition tasks. Each task is constructed by $K'$ ($0$$<$$ K'$$<$$K$) observed classes and the other $K$$-$$K'$ classes refer to the unseen class names that have appeared in the open vocabulary $\mathcal{V}$. 
    %\hspace{-0.2em}Afterwards, we select training images with the labels corresponding $K'$ observed classes to ensure a ratio of image-text training pairs with meaningful association, then randomly select the other training images to associate with the unseen classes. 
    Hence given each task $\mathcal{T}_i$ with $K$$-$$K'$ observed classes $\mathcal{V}_{i}$ and unseen classes $\mathcal{U}_{i}$, we define the task-specific energy function $E_{\phi}(X,H;\mathcal{T}_i)$ to capture the out-of-domain uncertainty:
    \vspace{-4pt}\begin{equation}\label{ef}
    	\begin{aligned}
    		E_{\phi}(X,H;\mathcal{T}_i) &=\hspace{-0.3em}\log \sum_{c\sim U_{i}} P_{\boldsymbol{\phi}}(X,H)[c]\\
    		=\hspace{-0.3em} \log \sum_{c\sim \mathcal{U}_{i}}&\frac{\exp\big(\frac{{\rm{sim}}(f(X), H(c;\mathcal{V}_i\cup \mathcal{U}_i))}{\gamma}\big)}{\sum_{i=1}^{K}\exp\big(\frac{{\rm{sim}}(f(X), H(c;\mathcal{V}_i\cup \mathcal{U}_i))}{\gamma}\big)},
    	\end{aligned}
    \end{equation}in which $\boldsymbol{x}$, $\boldsymbol{H}_{\phi}$ from Eq.\ref{mpl} is rewritten into $X$, $H$, respectivley, for representing random variables in the energy function in the range of the training data distribution. $E_{\phi}(X,H;\mathcal{T}_i)$ derived from $E_{\phi}(X,H)$ is suggested to meta-learn unseen concepts in $U_{i}$ across different tasks during training. Then, EMPL objective is defined as
     \begin{equation}\label{empl}
    	\begin{aligned}
    		\min_{\phi} \mathbb{E}_{\mathcal{T}_i} \bigg[&\underbrace{\mathbb{E}_{p(X,H|\mathbf{\mathcal{T}_i})}\Big[\sum_{c\sim V_{i}}-\log P_{\boldsymbol{\phi}}(X,H)[c]\Big]}_{\rm Generic \ prompt \ learning \ goal}\\& \ \ \ \ \ \ \ \ \underbrace{-\lambda\mathbb{E}_{p^{({\rm EBM})}_{\overline{\phi}}(X,H|\mathbf{\mathcal{T}_i})}\Big[E_{\phi}(X,H;\mathcal{T}_i)\Big]}_{\rm EBM \ uncertainty \ modeling}\bigg],
    	\end{aligned}
    \end{equation}where $p(X,H|\mathcal{T}_i)$ denotes the image-prompt training pairs extracted with their labels for achieving the maximum log-likihood prompt-tuning goal in terms of the task $\mathcal{T}_i$ (the first term); $p^{({\rm EBM})}_{\overline{\phi}}(X,H|\mathbf{\mathcal{T}_i})$ represents the image-prompt joint distribution derived from the task-specific energy function $E_{\phi}(X,H;\mathcal{T}_i)$. It is noteworthy that, $\overline{\phi}$ denotes the parameter frozen of the current iteration, hence $p^{({\rm EBM})}_{\overline{\phi}}(X,H|\mathbf{\mathcal{T}_i})$ is only available of generating image-prompt training pairs to compute the expectation for the second term. \hspace{-0.3em}The hyper-parameter $\lambda$ balances the strengths between in-domain image \hspace{-0.05em}recognition \hspace{-0.05em}and \hspace{-0.05em}open-vocabulary\hspace{-0.05em} concept \hspace{-0.05em}exploration. 
     
      \begin{figure}[t]
     	\center
     	\includegraphics[width = 0.95\columnwidth]{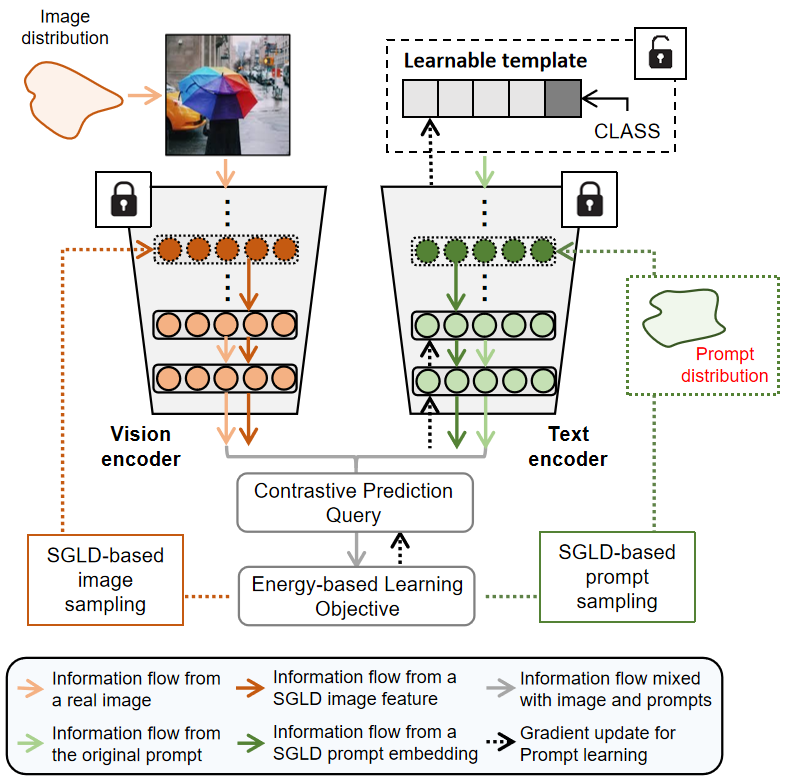}
     	\vspace{-0pt}
     	\caption{The paradigm of EMPL (best viewed in color). Briefly speaking, EMPL defines a prompt distribution based upon a EBM with the variables lying in the image feature and prompt embedding spaces. It categorizes an image with multiple prompts iteratively drawn from the EBM-based distribution by SGLD samplers. 
     	}
     	\vspace{-4pt}
     	\label{framework}
     \end{figure}
     
  { The energy function $E_{\phi}(X,H;\mathcal{T}_i)$ interacts with prompt learning by the second term, encouraging the low energy in terms of the image-prompt pairs extracted from $p(X,H|\mathbf{\mathcal{T}_i})$, in turn, stay high if the $p^{({\rm EBM})}_{\overline{\phi}}(X,H|\mathbf{\mathcal{T}_i})$ deviates from the marginal image-prompt distribution $p(X,H|\mathbf{\mathcal{T}_i})$.} The training process can be proved to endow the image-prompt pairs with a profound property:    
    	\begin{prop}\label{pro3}\vspace{-2pt}
    	Provided an arbitrary task $\mathcal{T}_{i}$ constructed to classify an image into either observed classes $V_i$ with avaiable training images or unseen classes $\mathcal{U}_{i}$ without unavailable training images, if $p(X,H|\mathbf{\mathcal{T}_i})$ denotes the marginal distribution of image-prompt pairs extracted from training data distribution, and $p_{\phi}(X,H|\mathcal{U}_i)=\sum_{c\sim \mathcal{U}_{i}} P_{\boldsymbol{\phi}}(X,H)[c]$ denotes the distribution of image-prompt contrastive prediction marginalized over all unseen classes in $\mathcal{U}_{i}$, the optimization objective of EMPL (Eq.\ref{empl}) encourages $p(X,H|\mathbf{\mathcal{T}_i})$ and $p_{\phi}(X,H|\mathcal{U}_i)$ negatively correlate with each other.
    	%		\begin{displaymath}
    		%		p_{\boldsymbol{\Phi}^\ast,\boldsymbol{\Theta}^\ast}(\boldsymbol{\hat{\boldsymbol{x}}}_j)=\int p_{\boldsymbol{\Phi}^\ast,\boldsymbol{\Theta}^\ast}(\boldsymbol{\hat{\boldsymbol{x}}}_j|\boldsymbol{x}_i)p_i(\boldsymbol{x}_i)d\boldsymbol{x}_i=p_{j}(\boldsymbol{\hat{x}}_j).
    		%		\end{displaymath}
    \end{prop}\vspace{-2pt}The theoretical result derived from uncertainty modeling \cite{wang2021energy}, enlightens us to understand the superiority of EMPL. In particular, when a well-trained EMPL model is provided with in-domain images, the energy-based prompt distribution is encouraged to assign a low contrasive score to any prompt without matching the images with correct classes since in-domain image-prompt pairs well match $p(X,H|\mathbf{\mathcal{T}_i})$ so that squeezes the value of $P_{\boldsymbol{\phi}}(X,H)[c]$ for all $c$ that falls within $\mathcal{U}_{i}$. With images drawn from the other domains or unseen classes, the image-prompt pairs are far from $p(X,H|\mathbf{\mathcal{T}_i})$, equivalently to increase $p_{\phi}(X,H|\mathcal{U}_i)$ for exploring the proper matching between the images and the unseen classes in $U_i$.

 %   enforces the in-domain prediction approach to the training marginal density and moreover, $\sum_{c\sim U_{i}}\log P_{\boldsymbol{\phi}}(X,H)[c]$ is negatively correlated with the training image-prompt distribution, 
    
 %   . In other words, given the unseen class name does not deliver any meaningful information to classify the samples drawn from the observed classes.  
    
 %   The energy-based prompts are optimized to produce high confidence to the training image-class pairs and give the word representing unseen classes with high energy to denote the uncertainty.
     
   % In terms of multi-prompt learning formulation, we first observe that 

    \textbf{SGLD Sampling and Prompting.} It is pivotally important of sampling image-prompt pairs from the energy-based distribution $p^{({\rm EBM})}_{{\phi}}(\cdot)$ since it does not only provide training instances for the second term in Eq.\ref{empl} but also generate $\boldsymbol{H}_{\boldsymbol{\phi}}$ to categorize $\boldsymbol{x}$ by the multi-prompting (Eq.\ref{ebm}). We employ Stochasitc Gradient Langevin Dynamics (SGLD) \cite{welling2011bayesian} to alternatively execute the sampling process\footnote{In terms of the meta-learning formulation in Eq.\ref{empl}, $E_{\phi}(\cdot,\cdot)$ is replaced by $E_{\phi}(\cdot,\cdot;\mathcal{T}_i)$ to denote the SGLD sampling process executed for the task.}: 
    \begin{equation}\label{sgld}
    	\begin{aligned}
    		\boldsymbol{x}^{t+1}&=\boldsymbol{x}^{t}-\frac{\alpha}{2}\frac{\partial E_{\phi}(\boldsymbol{x}^{t},\boldsymbol{h}^{t})}{\partial\boldsymbol{x}^{t}}+\sqrt{\alpha}\epsilon_1,& &\epsilon_1\sim\mathcal{N}(0;I),\\
    		\boldsymbol{h}^{t+1}&=\boldsymbol{h}^{t}-\frac{\alpha}{2}\frac{\partial E_{\phi}(\boldsymbol{x}^{t+1},\boldsymbol{h}^{t})}{\partial\boldsymbol{h}^{t}}+\sqrt{\alpha}\epsilon_2,& &\epsilon_2\sim\mathcal{N}(0;I),
    	\end{aligned}
    \end{equation}where $t$ and $\alpha$ denote the iteration and the step-size in the stochastic process; $\epsilon_1$ and $\epsilon_2$ are random noises drawn from a Gaussian distribution, respectively. The sampling process run in the feature space of $\boldsymbol{x}^{t+1}$ and the embedding space of $\boldsymbol{h}^{t+1}$. It significantly reduces the computational burden.   

    \textbf{Comparison with Other Prompt-Distribution Methods.} \hspace{-0.1em}Although previous multi-prompt learning efforts \cite{lu2022prompt,Alpher02}\hspace{-0.1em} have treated prompts as instances drawn from a distribution, our EMPL are predominant from some aspects. Concretely, the previous methods draw prompts from a specific type of density function or have to maintain a pre-defined context vector collection. Instead, EMPL defines the prompt distribution via a EBM derived from multi-prompt learning objective, where multiple prompts are dynamically drawn via executing a SGLD process, requiring little extra parameters beyond the base contexts. It prevents multi-prompt learning from the higher risk of overfitting. Besides, the EBM-based prompt distribution typically generates dynamic prompts conditioned on the visual feature. It resembles the spirit of CoCoOp distinct from other works.
        
%    \textbf{Theoretical Insight.}  More importantly, we theoretically demonstrate that when inquiring an unseen word, the open-vocabulary (OV) prediction learned by EMPL is \emph{negatively correlated} with the image distribution observed for prompt-tuning.   
        
%	Graph-Aligned Prompt Learning (GAPL) is proposed to overcome the shortcomings of existing multi-prompt learning across a broader range of label complexity and instance density. The overview has been shown in Figure 2. It is different from existing approaches with two techniques: \emph{Multi-prompt Meta-Net} and \emph{Graph-Aligned Cross-modal Prompting}. The former define how to generate multiple prompts to reflect visual aspects of different instances. The latter geminates the recent advance of discrete optimal transport \cite{bibid} to connect prompts and instances in a joint embedding space, upgrading prompt-based category prediction to many open-set concept learning downstream tasks.   

	\section{Experiments}
	In this section, we conduct comprehensive experiments to evaluate EMPL with diverse prompt learning approaches across three tasks, \emph{e.g.}, \emph{base-to-new generalization}, \emph{cross-domain generalization}, and \emph{cross-dataset transfer learning}. It provides the answer of our previous concern. %It exactly resembles the evaluation setup in CoCoOp \cite{Alpher02}.
	   
	   \begin{table}[t]
	   	\centering
	   	\setlength{\tabcolsep}{2.5pt}
	   	{\fontsize{7.2}{12}\selectfont
	   		\begin{tabular}{c c c c c c  c c c c c  }
	   			\hline
	   			\multicolumn{2}{c}{}&\multicolumn{2}{c}{{Single-prompt}}&&\multicolumn{4}{c}{Multi-prompt learning}\cr\cline{3-4}\cline{6-9}
	   			& \multirow{2}*{CLIP} & \multirow{2}*{\zlorange{CoOp}}  & \multirow{2}*{CoCoOp} &  &\multirow{2}*{\zlgreen{ProDA}}&\multirow{2}*{PLOT$^\ast$}  & \zlorange{CoOp} &\zlgreen{ProDA} \cr
	   			&&&&&&&\zlorange{(+EMPL)}&\zlgreen{(+EMPL)}\cr\hline
	   			Base &69.34&\zlorange{82.66}&80.47&&\zlgreen{81.56}&75.90&\zlorange{\textbf{82.73} (+0.07)}&\zlgreen{82 (+0.44)}  \\
	   			New &74.22 &\zlorange{63.22} &71.69&&\zlgreen{72.29} &67.6 &\zlorange{70.93 (+7.71)} &\zlgreen{\textbf{73.27} (+0.98)} \\
	   			H &71.69 &\zlorange{71.65} &75.83 &&\zlgreen{76.65} &71.8 &\zlorange{76.38 (+4.73)} &\zlgreen{\textbf{77.39} (+0.74)}  \\
	   			\hline
	   	\end{tabular}}\vspace{5pt}\caption{Comparison of single-prompt and multi-prompt learning baselines in the base-to-new generalization setting. $H$ (Harmonic mean \cite{xian2017zero}) measures the generalization trade-off. Different background colors indicate the corresponding group of abalating EMPL for CoOp and ProDA. }\label{t1}\vspace{-12pt}
	   \end{table} 
	\subsection{Experimental Setup}\vspace{-0pt}
    \textbf{Benchmarks.} The three tasks with fifteen datasets evaluate cross-modal prompt learners from different aspects. In terms of the base-to-new generalization and cross-dataset transfer setups, it takes ImageNet \cite{deng2009imagenet}, Caltech101 \cite{fei2004learning} for normal object recognition; SUN397 \cite{xiao2010sun} for
    scene recognition; UCF101 \cite{soomro2012ucf101} for action recognition;
    DTD \cite{cimpoi2014describing} for texture classification; EuroSAT \cite{helber2019eurosat} for satellite image classification; and OxfordPets \cite{parkhi2012cats},
    StanfordCars \cite{maji2013fine}, Flowers102 \cite{nilsback2008automated}, Food101 \cite{bossard2014food}, FGVCAircraft \cite{krause20133d} for fine-grained image recognition derived from diverse scenarios. 
    For the domain generalization, it trains models on ImageNet, then report the evaluation on ImageNetV2 \cite{recht2019imagenet}, ImageNet-Sketch \cite{wang2019learning}, ImageNet-A \cite{hendrycks2021natural}, and ImageNet-R \cite{hendrycks2021many}. The evaluation metric refer to the average accuracies and we additionally report the Harmonic mean \cite{xian2017zero} for base-to-new generalization, which is broadly regarded to judge the traded-off performances between the base and new classes. 
	
	\textbf{Baselines.} Beyond our methodology and  CLIP \cite{CLIP}, we consider CoOp \cite{CoOp}, CoCoOp \cite{CoCoOp}, ProDA \cite{lu2022prompt}, and PLOT \cite{chen2022prompt} for our comparison. Their backbones are typically derived from the open-source CLIP\footnote{\url{https://github.com/openai/CLIP}}, in which CoOp, CoCoOp are supposed to be the single-prompt learning baselines and ProDA, PLOT denote the multi-prompt learning baselines (We take PLOT$^\ast$ instead of PLOT to indicates its backbone distinct from other baselines). Note that EMPL is orthogonal to most existing prompt-tuning based methods and can be deployed to improve their strategies by energy-based multi-prompting. In this regard, our implementation for EMPL are derived from CoOp and ProDA\footnote{EMPL might also suit CoCoOp and PLOT in the spirit whereas it is prohibitively implemented by their open-source versions due to the heavy memory consumption for training.}, whereas their original objectives have been considered as the first term of Eq.\ref{empl} and their prompt generations are substituted by our SGLD-based sampler rather than their primitive strategies. We use EMPL(+CoOp) and EMPL(+ProDA) to represent their marriages, respectively.  
	
	\textbf{Implementation.} EMPL(+CoOp) and EMPL(+ProDA) are implemented with the public codes of CoOp and ProDA, in which we reformulate their learning objectives by introducing an energy-based function derived from the learnable contexts to specify SGLD-based samplers. The details of SGLD-based sampler and open-vocabulary meta-learning strategy refer to our Appendix.

	\subsection{Base-to-new Generalization}
	\textbf{Task setup.} It requires the prompt learner trained for few-shot generalization on the $10$ datasets with three different random
	seeds. Each dataset is divided into two disjoint subsets with base classes and new classes, where baseline models are trained with base classes via few-shot learning and evaluated on both base and new classes in the test dataset. For a fair comparison, we follow the dataset split and the number of shots in \cite{CoCoOp} during training.
	
	\begin{figure}[t]
		\center
		\includegraphics[width = 0.65\columnwidth]{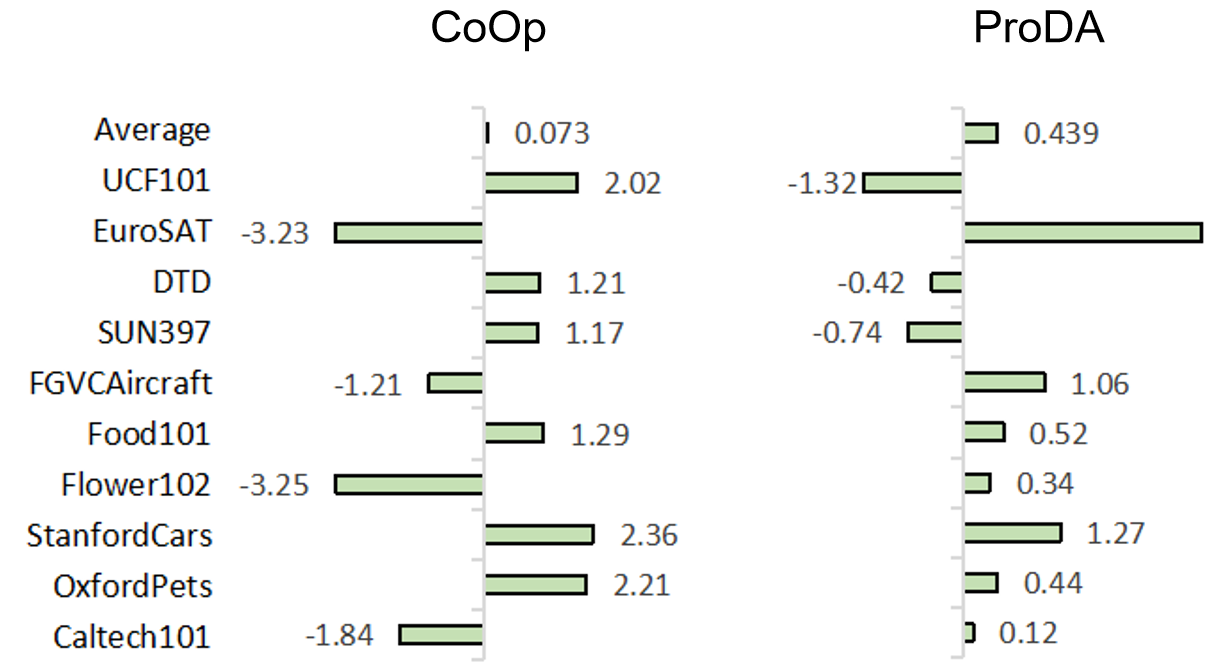}
		\vspace{-0pt}
		\caption{The performance change in base classes in 11 datasets. 
		}
		\vspace{-6pt}
		\label{base}
	\end{figure}
\textbf{Results.} Due to the space limitation, we report the average base-class and new-class accuracis along with their Harmonic mean over all datasets in Table.\ref{t1}, then providing the performance ablation for each dataset (Fig.\ref{base},\ref{new}). As demonstrated in Table.\ref{t1}, CoOp is a competitive rival in Base-class generalization with regard to its outperformance compared with other baselines beyond the range of our EBML. But its superiority remains a doubt of overfitting because its accuracy rapidly drops while coming to the unseen classes. The performance discrepancy could be greatly mitigated when CoOp takes the prompts generated by EMPL. The marriage leads to +7.71 performance gain on the new classes, driving the Harmonic mean to 76.38 that sufficiently defeats all single-prompt learning baselines. CoCoOp is famous as a complementary strategy to CoOp, while its pipeline suffers from the low inference efficiency due to its instance-specific prompt scheme demanding an independent forward pass for each prompt. With this regard, CoCoOp hardly becomes, or combines with a multi-prompt strategy and thus, inevitably fall behind all ProDA variants. Notice that, EMPL endows ProDA with visually-encoded information conducted by the SGLD-based prompting scheme (Eq.\ref{sgld}). It results in the uppermost trade off in the base-to-new generalization .
	
	We further ablate EMPL in the CoOp and ProDA across all datasets. As shown in Fig.\ref{base}, EMPL benefits the majority of tasks ($7$ of $10$ in CoOp and $6$ of $10$ in ProDA) with moderate margins whereas also produces unexpected negative effects to CoOp and ProDA for the minority. It probably owes to the conservative tendency to the observed classes for open-vocarbulary meta-learning, in which the energy function aims to maintain the exploitation-exploration balance between in-domain classes and out-of-domain classes (Proposition.\ref{pro3}). Notwithstanding, the base-class predominances of CoOp and ProDA go on without sacrificing the new-class generalizability in Fig.\ref{new}.       
		
	\begin{figure}[t]
	\center
	\includegraphics[width = 0.9\columnwidth]{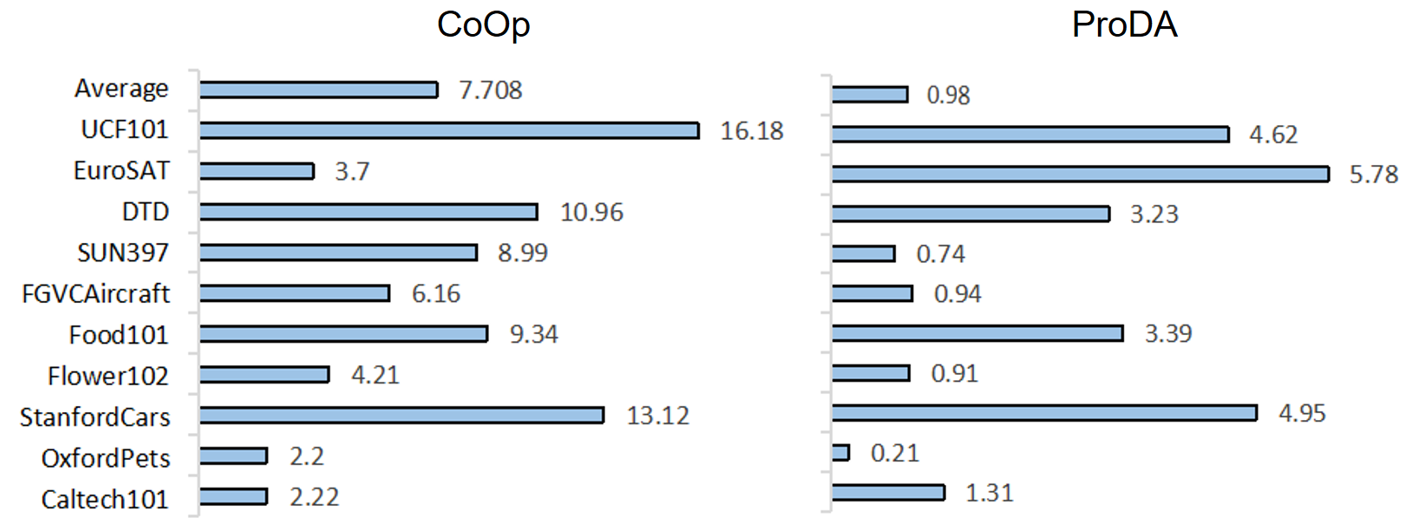}
	\vspace{-0pt}
	\caption{The performance change in new classes in 11 datasets. 
	}
	\vspace{-0pt}
	\label{new}
\end{figure}
\begin{table}[]
	\centering
	\setlength{\tabcolsep}{2.5pt}
	{\fontsize{8.4}{12}\selectfont
		\begin{tabular}{c c c c c c  c c ccc  }
			\hline
			\multicolumn{1}{c}{}&\multicolumn{2}{c}{{Source}}&\multicolumn{4}{c}{Target}\cr\cline{2-2}\cline{4-7}
			&ImageNet&&-V2&-Sketch&-A&-R \cr
			CLIP &66.73&&60.83&{46.15}&47.77&73.96 \\
			\rowcolor{SeashellO}	CoOp &\textbf{71.51}&  &64.20&47.99 &49.71 &75.21\\
			\rowcolor{SeashellO}		CoOp(+EMPL) &70.89 $\downarrow$&  &64.91 $\uparrow$&48.64 $\uparrow$ &{51.27} $\uparrow$ &76.01 $\uparrow$\\
			CoCoOp &71.02  &&64.07 &\textbf{48.75} &50.63  &76.18 \\
			\rowcolor{SeashellG}ProDA &71.41 &&65.14  &{46.78} &51.62& 75.67\\ \rowcolor{SeashellG}ProDA(+EMPL) &71.17 $\downarrow$ &&64.79 $\downarrow$ &{48.42} $\uparrow$ &\textbf{52.35} $\uparrow$&76.84 $\uparrow$\\
			\hline
	\end{tabular}}\vspace{5pt}\caption{Comparison of single-prompt and multi-prompt learning baselines for cross-domain generalization.(best viewed in color) }\label{t3}\vspace{-16pt}
\end{table} 
	
	\subsection{Cross-domain Generalization}
	
	\textbf{Task setup.} Distinct from the base-to-new setup, cross-domain generalization attempts to examine the baselines in terms of their resiliences against domain shift and adversarial robustness: models trained in ImageNet are evaluated on its four target dataset variants for different purposes. 
		\begin{table*}[]
		\centering
		\setlength{\tabcolsep}{2.5pt}
		{\fontsize{7.8}{11}\selectfont
			\begin{tabular}{c c c c c c  c c ccc cc cc}
				\hline
				\multicolumn{1}{c}{}&\multicolumn{2}{c}{{Source}}&\multicolumn{10}{c}{Target}\cr\cline{2-2}\cline{4-13}
				&ImageNet&&Caltech101 &OxfordPets&StanfordCars&Flower102&Food101&FGVCAircraft&SUN397&DTD&EuroSAT&UCF101&Avg \cr
				\rowcolor{SeashellO}	CoOp &\textbf{71.51}&  &93.7&89.14 &64.51 &68.71&85.35&18.47&64.15&41.92&46.39&66.55&63.88\\
				\rowcolor{SeashellO}		CoOp(+EMPL) &70.89 &  &94.16 &90.21  &{65.29}  &71.52 &86.21&23.16&67.13&46.93&\textbf{47.34}&68.07&66.49\\
				CoCoOp &71.02  &&94.43&90.14&65.32&\textbf{71.88}&86.06 &22.94&67.36&45.73 &45.37  &68.21&65.74 \\
				\rowcolor{SeashellG}ProDA &71.41&&\textbf{94.65}&90.22&64.81&70.69&85.57&22.23&\textbf{68.23}&43.33&45.78&67.86&65.89\\ \rowcolor{SeashellG}ProDA(+EMPL) &71.17&&94.63&\textbf{91.24}&\textbf{65.67}&71.76&\textbf{86.29}&\textbf{23.97}&67.98&\textbf{47.21}&46.87&\textbf{68.44}&\textbf{66.81} \\
				\hline
		\end{tabular}}\vspace{0pt}\caption{Comparison of single-prompt and multi-prompt learning baselines for cross-dataset transfer.(best viewed in color) }\label{t4}\vspace{-6pt}
	\end{table*} 	
	
	\textbf{Results.} As reported in Table.\ref{t3}, our EMPL enhances the cross-domain accuracis of CoOp and ProDA in seven out of eight situations with diverse type of distribution shifts. It fails in the cases with the source data for training (ImageNet for CoOp) or with a mild distribution shift (ImageNet-V2 for ProDA). In terms of the cases containing significant visual difference (ImageNet-Sketch), out-of-distribtuion shift (ImageNet-R) and natural adversarial noises (ImageNet-A), our EMPL consistently emerged victorious for the generalization across domains.       
	
	\vspace{-0pt}\subsection{Cross-dataset Transfer}\vspace{-0pt}
	
	\textbf{Task setup.} We finally evaluate the baselines in the more challenging cross-dataset transfer setups, whose fundamentals are allowed to totally change across datasets (different tasks across different domains). In this case, the baseline prompt contexts are trained on ImageNet, then, required to access on the other target datasets with distinct knowledge. 
	
%	This experiment tries to determine how effectively our	methods generalizes transfer beyond the scope of a single dataset

	\textbf{Results.} As demonstrated in Table.\ref{t4}, all baselines perform similarly in the source training set while behave differently across diverse target datasets, and in most cases, the multi-prompt learners outperform the single-prompt learners. In particular, EMPL variants outperform the other baselines in seven datasets and more importantly, they have significantly benefited their basic models with the accuracy increases in $18$ of $20$ transfer scenarios. More typically on the target datasets such as FGVCAircraft, SUN397, and DTD, EMPL raised their accuracies more than $3$$\%$ and besides, it does not introudce any extra parameters to acheive this goal. On the other side, we also observe that EMPL results were not the state of the art (ImageNet, Caltech101, Flower102, and SUN397). The ImageNet case is similarly explained as what happened in Table.2-3, \emph{i.e.}, the boost by EMPL is largely due to the new-class generalization. So testing EMPL on the source data may deemphasize this merit. As to Caltech101, Flower 102, and SUN397, it is observed that EMPL just slightly underperform the state-of-the-art models, \emph{e.g.}, ProDA is 94.65 yet EMPL(+ProDA) is 94.63. As the number of promptgs was fixed to 8, their performances might be further improved by prompt augmentation.
	
	%we relax this computation-burden contraint to observe whether the EMPL (+ProDA) could be further improved by grid-searching the optimal number of prompts in (8,16,32,64). Fig.\ref{fig1}(Rebuttal) illustrates that we can adapt EMPL to a dataset with the optimal amount of prompt embeddings, so it may exceed the previous SOTAs 

		\begin{figure}[t]
		\center
		\includegraphics[width = 1\columnwidth]{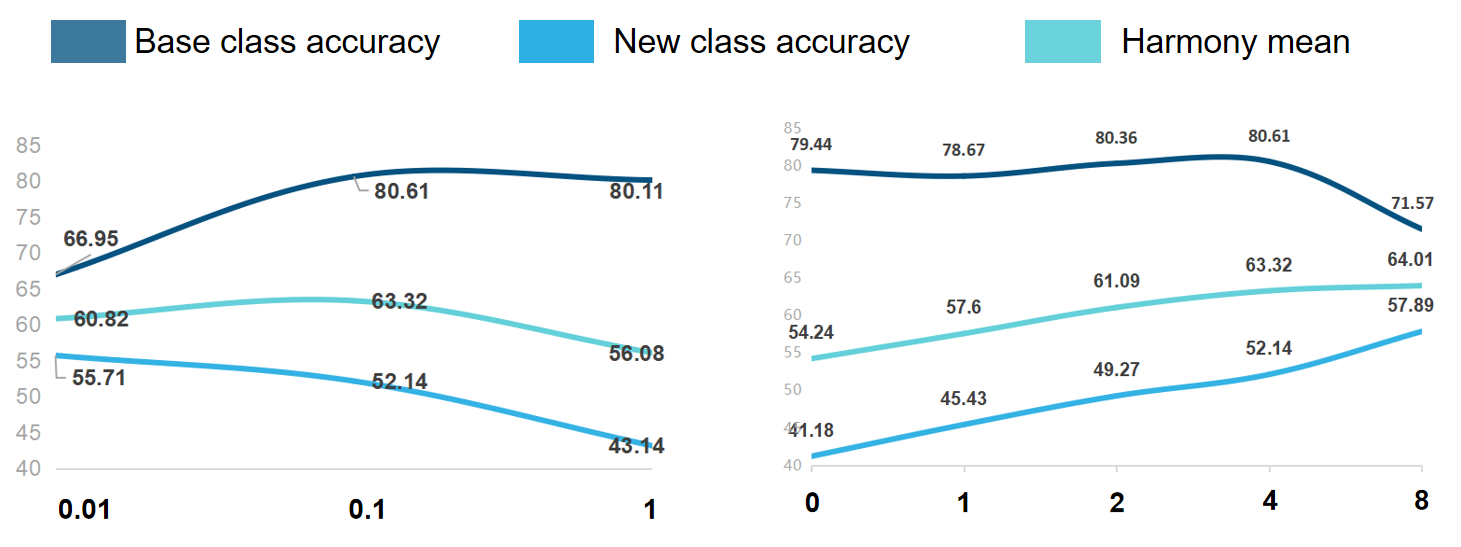}
		\vspace{-4pt}
		\caption{The trending curves when changing the value of $\lambda$ and the number of new words for training (best viewed in color). 
		}
		\vspace{-4pt}
		\label{lk}
	\end{figure}
	\vspace{-0pt}\subsection{Analysis}\vspace{-0pt}
		\textbf{The number of prompts.} The size of prompts used for each prompting inference sufficiently affects the final performance for arbitrary prompt distribution methods. So we provide the ablation for evaluating EMPL with the prompt number used for training. We evaluate EMPL variant model derived from CoOp based on DTD dataset, then, observe the change of the base-class, new-class accuracies, and their Harmonic means. As reported in Fig.\ref{np}, the performance of EMPL could be increased by drawing more prompt embeddings from the energy-based distribution. 
		
	\textbf{Hyperparameters $\lambda$ and $K$$-$$K'$.} The open-vocabulary meta-learning objective (Eq.\ref{empl}) plays a key role of achieving the trade-off balance in base-to-new generalization. The implemenation is typically related with the hyperparameters $\lambda$ and $K$$-$$K'$: the former determines the sensitivity to explore visual patterns with new-word prompts; the latter controls the ratio of how many new words would appear per training batch. We follow the ablation setup above and then change the value of $\lambda$ in the range $\{0.01$,$0.1$,$1\}$, then, taking the same evaluation setup by changing the number of unseen words per training batch in the range $\{1$,$2$,$4$,$8\}$. As we have observed in Fig.\ref{lk}, $\lambda$ typically trades off the in-domain and out-of-domain results where its larger value implies the objective with more attention to explore the uncertainty. It leads to the rise of new-class generalization, vice and versa. $K$$-$$K'$ is also related with the new-class generalization performance: increasing the number of new words leads to the improvement while it rapidly converges to the bottleneck.
		\begin{figure}[t]
		\center
		\includegraphics[width = 0.45\columnwidth]{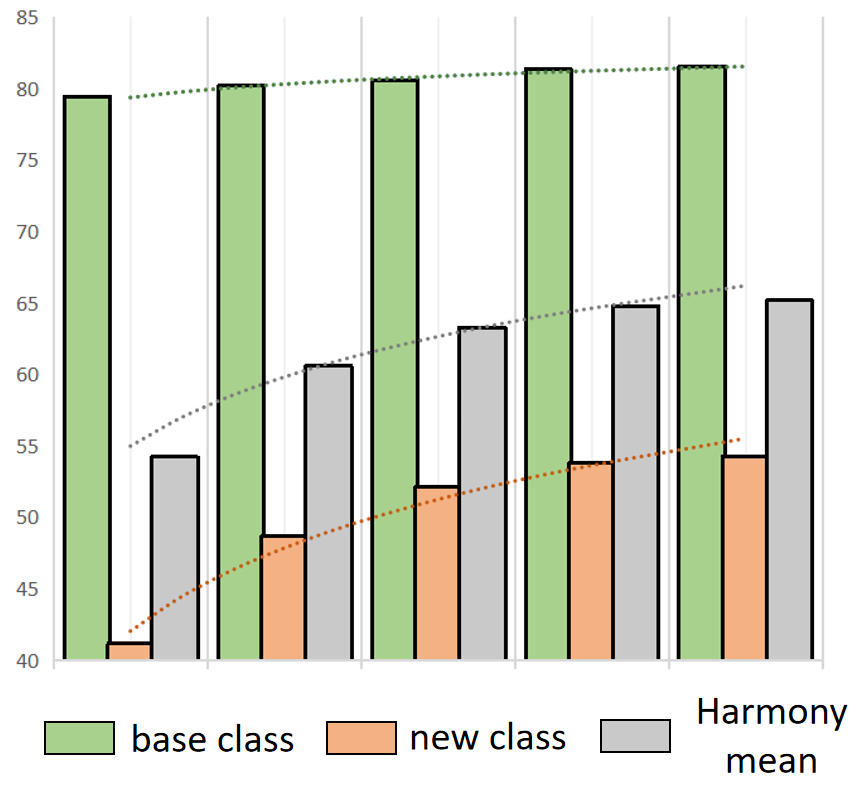}
		\vspace{-0pt}
		\caption{The statistic and trending curves when changing the number of prompt samples for training (best viewed in color). 
		}
		\vspace{-10pt}
		\label{np}
	\end{figure}

	\begin{table}[h]
	\centering
	\setlength{\tabcolsep}{2.5pt}
	{\fontsize{8.4}{12}\selectfont
		\begin{tabular}{c c c c c ccccc  c c ccc  }
			\hline
			Transformer	&\multicolumn{3}{c}{{Performance}}&&\multicolumn{1}{c}{Computation burden}\cr\cline{2-4}\cline{6-6}
			layers	&Base&	New&	H&&Sec/iter \cr\hline
			2	 &80.61&52.14&63.32&&0.048 \\
			3 &80.71& 52.2 &63.39&&0.072\\
			6 &79.41&53.31  &63.79&&0.143\\
			9		&77.28&51.24  &61.62&&0.278\\
			\hline
	\end{tabular}}\vspace{5pt}\caption{The trade-off between performance (the base-to-new generalization setup in DTD) and computation burden.}\label{b}\vspace{-4pt}
\end{table} 
	\textbf{Positions for SGLD-based sampling.} EMPL's training and prompting rely on SGLD that runs in the embedding space. To justify the concern of backward sampling computation, we ablate EMPL (+CoOp) variants with the SGLD sampler applied to different
	positions in the text encoder and take sec / iter to measure how long it takes to generate a prompt embedding with a single RTX 3090 GPU. Table.\ref{b}
	shows that applying the SGLD sampling to the low-level space
	incurs huge computation overhead without obvious performance bonuses. It encourages us to take the two-layer backward pass to generate prompts across all experiments.

%	\subsection{Zero-shot Task Generalization for Multi-label Image Recognition}

	\subsection{Vision-Language Information Retrievel} 

	 We finally provide the empirical study with respect to image-text retrieval tasks on MSCOCO and Flickr30K. We employed Karpathy split to separate MSCOCO into 113/5K/5K and Flickr30K into the amouts of 29,000 / 1,000 /1,000 for training / validation / test sets, respectively. We further construct the few-shot subsets for prompt tuning, with 0.5\% and 1\% instances drawn from their training sets, respectively. Given this, we train the prompt learners with these subsets, then evaluate the prompt learners' performance on their corresponding test sets using Recall at 1 (R@1) as our evaluation metric.We focus on the evaluation to CLIP, CoOp, CoCoOp and EMPL (+CoOp), where CLIP did not join prompt tuning and the other captions took as images' class labels. In Table.\ref{tr}, we observe that the CoOp-based models trained with 0.5\% data in MSCOCO and Flick30K both suffer from the overfitting compared with CLIP. Encoding visual information by CoCoOp helps to alleviate, but failed to solve it in Flickr30K. In contrast, EMPL prevented CoOp from overfitting to the scarce training subsets to achieve the optimal results. With 1\% training data, EMPL significantly improves CoOp, \emph{e.g.}, \textbf{+3.37} for MSCOCO (1\%) and \textbf{+2.92} for Flickr30K (1\%), which outperformed the other baselines.  
	 
	 	\begin{table}[t]
	 	\centering
	 	\setlength{\tabcolsep}{2.5pt}
	 	{\fontsize{7.6}{12}\selectfont
	 		\vspace{-3pt}	\begin{tabular}{c c c c c c  c c c c c  }
	 			\hline
	 			&{0\%}&&\multicolumn{4}{c}{{0.5\%}}&\multicolumn{3}{c}{1\%}\cr\cline{2-2}\cline{4-6}\cline{8-10}
	 			& {CLIP} && {{CoOp}}  & {CoCoOp} &{{EMPL}}&&{CoOp}  & {CoCoOp} &{EMPL}\cr\hline
	 			MSCOCO &53.35&&{53.10}{\color{red}$\downarrow$}&54.50{\color{green}$\uparrow$}&\textbf{55.45}{\color{green}$\uparrow$}&&53.58{\color{green}$\uparrow$}&{56.40}{\color{green}$\uparrow$}&\textbf{56.85}{\color{green}$\uparrow$}  \\
	 			Flickr30K &83.06 &&{81.90}{\color{red}$\downarrow$} &82.80{\color{red}$\downarrow$}&\textbf{83.94}{\color{green}$\uparrow$}&&82.71{\color{red}$\downarrow$} &{84.50}{\color{green}$\uparrow$} &\textbf{85.63}{\color{green}$\uparrow$} \\
	 			\hline
	 	\end{tabular}}\vspace{-0pt}\caption{Image-text retrieval results on MSCOCO and Flick30K. }\label{tr}\vspace{-12pt}
	 \end{table}
	\section{Conclusion, Limitation, and Future Work}
	In this paper, we have proposed a systematic overview to vision-language multi-prompt learning. In the discussion scope of CLIP, we revealed why multi-prompt learning strategies can improve cross-modal transferrability: (1) multi-prompt learning empirically reduces the modality gap with prompt augmentation and (2). single prompt learning provably suffers from non-identifiability issue while augmenting the prompt may alleviate. Given this obervations, we propose a new energy-based multi-prompt learning (EMPL) approach to improve the open-vocabulary generalization capability with regards to uncertainty modeling. Our EMPL does not require any extra parameter introduced for CLIP, while its superiority has been theoretically and empirically supported by thorough experiments. 
		%The parameter efficiency of our Multi-prompt Meta-Net does not imply a low GPU memory consumption. Similar with . Instead of making prompts with all instances, we pay attention to the most representative feature subset $F$$\subset$$\{f_{1}, \cdots, f_{n_{\boldsymbol{x}}}\}$ for prompt generation. The subset $F$ is obtained by solving a contrained submodular maximization objective \cite{bibid} \footnote{See more in our Appendix.C} for both training and testing.
		
		The drawback of EMPL mainly comes from its time cost for the prompt embedding inference . According to our device for training, we paid the triple time cost more than the CoOp original version and even more in terms of ProDA-based EMPL variants. For each inference for testing, we are encouraged to take double prompt embeddings compared with training phase to increase the performance, where we take the most certain class as our prediction results.   
		
According to our discssion, it would be several promising treads in the future for multi-prompt learning. First, we only raise the first theoretic concern to multi-prompt learning since we have discovered the occurence of cross-modal non-identifiability behind single-prompt learner. Whereas why and how multi-prompt learners work out, still requiring for more sophisticated analytical studies with respect to learnability and optimization theories. Second, the prompt diversity is the key why multi-prompt learners outperform single-prompt learners. Therefore the research on the topic related with the prompt diversity is also inspiring.  Finally, multi-prompt learning were always proposed as a time-consuimg approaches since they have to infer their prompts multiple times to predict each image. Reducing the inference cost would be a crucial issue in this field.  

\section{Ackonwledgemet}

This work was supported in part by National Key R\&D Program of China under Grant No.2021ZD0111601; in part by National Natural Science Foundation of China (NSFC) under Grant No.61836012, U21A20470, 62206110, and 62077028; the Science and Technology Planning Project of Guangdong (No.2020ZDZX3013), the Science and Technology Planning Project of Guangzhou (No.202206030007) and the Opening Project of Key Laboratory of Safety of Intelligent Robots for State Market Regulation
(No.GQI-KFKT202205); in part by GuangDong Basic and Applied Basic Research Foundation under Grant No.2023A1515012845 and 2023A1515011374. Liang Lin is also leading the GuangDong Province Key Laboratory of Information Security Technology.

%	You must include your signed IEEE copyright release form when you submit your finished paper.	We MUST have this form before your paper can be published in the proceedings.
	
%	Please direct any questions to the production editor in charge of these proceedings at the IEEE Computer Society Press:
%	\url{https://www.computer.org/about/contact}.

	%%%%%%%%% REFERENCES

    {\small
    	\bibliographystyle{ieee_fullname}
    	\bibliography{egbib}
    }\clearpage
\end{document}